\documentclass[conference]{IEEEtran}
\IEEEoverridecommandlockouts
\usepackage{cite}
\usepackage{amsmath,amssymb,amsfonts}
\usepackage{algorithmic}
\usepackage{graphicx}
\usepackage{textcomp}
\usepackage{xcolor}
\def\BibTeX{{\rm B\kern-.05em{\sc i\kern-.025em b}\kern-.08em
    T\kern-.1667em\lower.7ex\hbox{E}\kern-.125emX}}
    
\usepackage{booktabs}

\newcommand{\R}{\mathbb{R}}
\newcommand{\N}{\mathcal{N}}

\newcommand{\model}{GRN }

\newcommand{\A}{\mathbf{A}}

\renewcommand{\u}{\mathbf{u}}
\newcommand{\x}{\mathbf{x}}
\newcommand{\y}{\mathbf{y}}
\newcommand{\W}{\mathbf{W}}
\newcommand{\V}{\mathbf{V}}

\newcommand{\U}{\mathbf{U}}
\newcommand{\X}{\mathbf{X}}

\begin{document}

\title{Ring Reservoir Neural Networks for Graphs}

\author{
\IEEEauthorblockN{Claudio Gallicchio}
\IEEEauthorblockA{\textit{Department of Computer Science} \\
\textit{University of Pisa}\\
Pisa, Italy \\
gallicch@di.unipi.it}
\and
\IEEEauthorblockN{Alessio Micheli}
\IEEEauthorblockA{\textit{Department of Computer Science} \\
\textit{University of Pisa}\\
Pisa, Italy \\
micheli@di.unipi.it}
}

\maketitle

\begin{abstract}
Machine Learning for graphs is nowadays a research topic of consolidated relevance.
Common approaches in the field typically resort to complex deep neural network architectures and demanding training algorithms, highlighting the need for more efficient solutions. 
The class of Reservoir Computing (RC) models can play an important role in this context, enabling to develop fruitful graph embeddings through untrained recursive architectures.
In this paper, we study progressive simplifications to the design strategy of RC neural networks for graphs. Our core proposal is based on shaping the organization of the hidden neurons to follow a ring topology. Experimental results on graph classification tasks indicate that ring-reservoirs architectures enable particularly effective network configurations, showing consistent advantages in terms of predictive performance. 
\end{abstract}



\section{Introduction}
Graph processing is assuming a central role in the development of the basic  Machine Learning (ML) research \cite{tutorial2020-arxiv}. Indeed, the possibility to
consider  the relationships among the singular samples  in form of graphs and
networks allows to overcome the limitations of  models for  flat domains, i.e. with 
 fixed-size sample representations (fixed-length vectors) for the input data. 
As such, the direct treatment of this kind of data offers a remarkable opportunity to extend the possibility of successful application domains in fields that range from the processing of language structures or molecular data to the analysis of social or biological networks (just to mention some noteworthy examples).

The are many trends in the area of processing graph data in ML with a historical root in models for the adaptive processing for hierarchical data (rooted trees) \cite{Sperduti1997,Frasconi1998, Hammer-book}, with pioneering applications to the Cheminformatics domains \cite{Appl-Int2000, JChem2001}, which have been  progressively extended to a family of models (and studies) based on the recursive approach. This family  includes a wide range of approaches, 
from the  unsupervised  \cite{Hammer-Neural-Networks-2004} and generative \cite{bhtmm} areas, up to the  extension  to directed acyclic graphs \cite{Micheli2004contextual}.
All this models have been characterized by a recursive definition of the states (embedding) associated to each node of the input structures that naturally extends the state transition system of a recurrent neural network (RNN) for sequential data. 

The direct processing of general directed/undirected and cyclic/acyclic graphs have been introduced following two basic approaches. The first one is based on a direct extension of the recursive neural networks, introduced  by the Graph Neural Network (GNN)\cite{scarselli2009graph}   and then for the Reservoir Computing framework by the Graph Echo State Network (GESN) \cite{gallicchio2010graph}. Such approach uses a recursive model where the cyclic dependencies among states of the recursive transition system are allowed and treated by imposing constraints on the resulting dynamical system (resorting to a contractive dynamics to assure the convergence to a fixed-point representation for the state of each vertex of a graph).   As a result, the fixed point of the recursive/dynamical system is exploited to represent (or {\it embed}) the input graphs. Once the states have been computed for 
all  graph vertices, iterating the state transition function until convergence,  the graph embedding values can be  projected to the model output (readout), which is implemented as a standard layer of trained neural units. The diffusion of the context for each node of the graph is also guided by the iterations used to reach the stable point of the state values.

Differently from the recursive models, the other line for graphs (coeval with the recursive approaches)  
is to exploit the idea of stacking 
feed-forward layers of neural units to manage the mutual dependencies among state values that can occur in cyclic and/or undirected graphs. The multi-layer construction allows the context information to be diffused for the state of each input node in a compositional way.   This line of research, based on a spatial approache, has been further  developed  under the general term of convolutional neural network (CNN) for graphs  by many authors after the 2015 \cite{zhang2018end,tran2018filter,atwood2016diffusion,niepert2016learning,xu2018powerful}.

Finally, a further 
class of approaches that deserved to be mentioned (and that we will consider in our experimental part) is given by kernel methods for graphs
%
\cite{shervashidze2009efficient,yanardag2015deep,vishwanathan2010graph,neumann2016propagation,shervashidze2011weisfeiler}. 

Since the processing of complex types of data comes at the cost of a high computational demand, another trend concerns the efficiency issue. For the case of sequences and trees, the Reservoir Computing (RC) paradigm
\cite{Lukosevicius2009}, in particular in the form of the   so-called Echo State Network (ESN) \cite{Jaeger2004},  provides an approach for the efficient modeling of recurrent/recursive models \cite{Gallicchio2013tree}. Such paradigm is based on the use of fixed (randomized) values of the recurrent weights under stability conditions of the dynamical system (Echo State Property - ESP) \cite{Jaeger2004},
while the output units (readout)  are  the only trained part of the model.
This line has been extended both to graph processing (i.e. the already mentioned GESN) and to deep architectures for trees \cite{INF-SCIENCE} and graphs \cite{gallicchio2020aaai}.

In this paper we follow the lines for efficient processing through RC approaches, with the aim to investigate on the effect of the neural network topology for processing graphs data.
In particular, the proposed  approach 
exploits the study on “minimum complexity” ESN for sequential data introduced in \cite{rodan2010minimum}, in which the connections between the recurrent neurons of architecture follow a specific (ring shaped) topology and the stochastic elements (randomization in the  recurrent weights initialization) are progressively eliminated, 
acting in accordance with  specific deterministic rules. As a result, each instance of the model is fully described only by the value of a few hyper-parameters.
The extension to graph domains lead us to introduce a new 
approach 
for graph processing by RC
neural networks 
based on ring architectures.

Beside the investigation end, our proposal aims to introduce practical advantages in terms of a rational and  simplified setting of the  GESN, allowing also 
the user to increase the number of trials during the model cross-validation phase without increasing the computation time, or, in other words, to extend  the exploration of the hyper-parameters space, which, in turns, increases the possibility to refine the application performance.

The rest of this paper is structured as follows.
In Section \ref{sec.rcgraph} we give the basics of 
RC 
for graph processing. Then, in Section \ref{sec.ringrc} we introduce the new proposed architectures, while in Section \ref{sec.experiments} we analyze the results on different benchmarks at the state-of-the-art in the field of neural networks for graphs, also analyzing the effect of the architecture with respect to the number of units and the computation resource. Finally, in Section \ref{sec.conclusions}   we draw our conclusions.

\section{Reservoir Computing for Graphs}
\label{sec.rcgraph}

\noindent
\textbf{Preliminaries on graphs.}
In this paper we focus on the problem of classifying undirected graph structures.
A graph $g$ is represented by the couple $g = (V_g, E_g)$, where $V_g$ denotes the set of vertices and $E_g$ is the set of edges. The number of vertices in $g$ is  indicated by $N_g$. The connectivity structure among the vertices of $g$ is compactly represented by its adjacency matrix $\A_g$, which is a square (in our case symmetric) $N_g \times N_g$ matrix whose $(i,j)$ element is always $0$ unless there is an edge connecting vertex $i$ to vertex $j$. The neighborhood of each vertex $v \in V_g$ is the set of vertices that are adjacent to $v$, denoted as $\N(v) = \{v' \in \V_g | (v,v') \in E_g\}$. The maximum among the sizes of the neighborhoods defines the degree $k$. We consider vertex-labelled graphs, where an input label (i.e. a feature vector) is assigned to each vertex. Here we use the notation $\u(v)$ to indicate the input label attached to vertex $v$. To ease the notation, in the the rest of the paper we drop the subscript $g$ whenever the reference to the graph in question is unambiguous.

\noindent
\textbf{Graph Echo State Networks.}
Reservoir Computing (RC) \cite{Lukosevicius2009} is a popular design paradigm for efficiently trained dynamical neural models, extended to process graph structures with the introduction of the Graph Echo State Network (GESN) in \cite{gallicchio2010graph}. 

Architecturally, a GESN  is composed by a hidden layer of recursive non-linear neurons, the \emph{reservoir}, that implements the graph embedding process, followed by a simple linear feed-forward readout layer that computes the output.
Crucially, and differently from conventional recurrent graph neural network approaches \cite{scarselli2009graph}, the weights on the hidden (reservoir) layer's connections are left untrained after initialization, and only the parameters of the readout are subject to training. 

\begin{figure*}[htbp]
\centering
\includegraphics[width=1\textwidth]{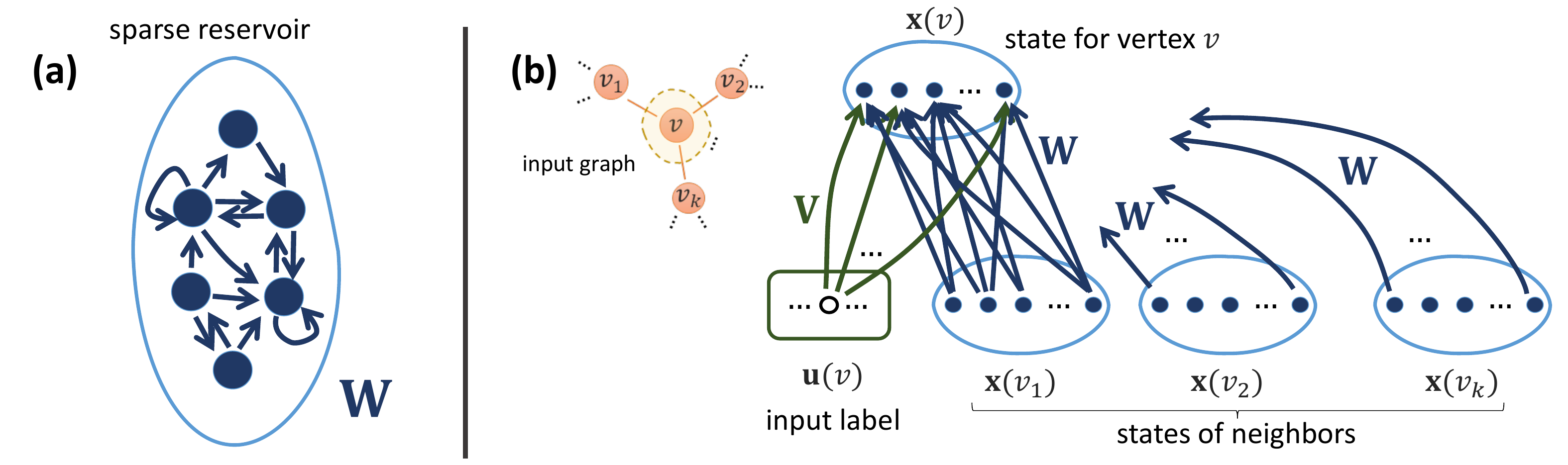}
\caption{GESN: sparse reservoir recurrent structure (a), and its application to graph encoding (b).}
\label{fig.GESN}
\end{figure*}

We use $N_I$, $N_H$ and $N_O$ to respectively indicate the number of input, hidden (i.e., reservoir) and readout units.
The reservoir's neurons encode each vertex $v$ of an input graph into a state representation (i.e., an internal neural embedding) by means of a state transition function defined as follows\footnote{We drop the bias terms in the equations for the ease of notation.}:
\begin{equation}
\label{eq.reservoir}
\x(v) = \tanh\big(\V \, \u(v) + \sum_{v' \in \N(v)}\W \, \x(v')\big),
\end{equation}
where $\x(v) \in \R^{N_R}$ is the state computed for vertex $v$, $\u(v) \in \R^{N_I}$ is the input label (i.e., the input feature vector) attached to vertex $v$, 
$\V \in \R^{N_H \times N_I}$ is the input weight matrix, $\W \in \R^{N_H \times N_H}$ is the reservoir recurrent weight matrix, and $\tanh$ indicates the element-wise application of the hyperbolic tangent non-linearity. As described by \eqref{eq.reservoir}, the state for each vertex $v$ depends on both its input features, modulated by the weights in $\V$, and on the set of states computed for the vertices in the neighborhood of $v$, modulated by the weights in $\W$.
In line with RC for time-series, the reservoir neurons are sparsely and randomly connected among each other. This implies that $\W$ is instantiated as a sparse matrix, hence speeding-up the state transition computation, which scales linearly with the number of reservoir neurons, with the degree and with the number of vertices \cite{gallicchio2020aaai}.
The reservoir recurrent structure and the state transition computation (applied to a vertex in the input graph) are graphically illustrated in Fig.~\ref{fig.GESN}.

The operation of the reservoir on an entire input graph can be collectively represented in the 
(more compact) form:
\begin{equation}
\label{eq.encoding}
\X = \tanh(\V \U + \W \X \A),
\end{equation}
where $\U \in \R^{N_I \times N}$ and $\X \in \R^{N_H \times N}$ column-wise collect respectively the input labels and states for all the $N$ vertices in the given input graph, and $\A$ is the adjacency matrix. Interestingly, \eqref{eq.encoding} can be studied as the function driving the dynamics of an input-driven non-linear dynamical system, where the role of exciting (external) input information is played by both $\U$ and $\A$ (that depend on the graph to which the reservoir system is applied). Existence and uniqueness of solutions of \eqref{eq.encoding}, i.e. of the graph embedding computed by the reservoir, can be guaranteed by imposing a stability property named \emph{Graph Embedding Stability} (GES) \cite{gallicchio2020aaai}. A necessary condition for the GES consists in scaling the effective spectral radius, i.e. $\rho = \rho(\W) \, k$, to a value smaller than 1, where $k$ denotes the degree of the set of graphs under consideration and $\rho(\W)$ is the largest eigenvalue of $\W$ in modulus \cite{gallicchio2020aaai}. The reservoir parameters in matrix $\W$ can  then be randomly initialized from a uniform distribution over $[-1,1]$, and then re-scaled to meet the $\rho < 1$ condition. Analogously, weights in the input matrix $\V$ are randomly initialized from a uniform distribution over $[-\omega, \omega]$, where $\omega$ is an input-scaling parameter that determines the strength of the input influence in driving the reservoir dynamics. 
The values of $\rho$ and $\omega$ are treated as hyper-parameters. Given an input graph, \eqref{eq.encoding} is then iterated until convergence\footnote{Typically, the reservoir is initialized to a zero state for each vertex in the graph, although any initial condition in the state space would be just as good due to the uniqueness of the solution of \eqref{eq.encoding}.} to the unique fixed point of the corresponding dynamical system. Such a fixed point in the reservoir state space depends on the specific input graph (through matrices $\U$ and $\A$) and represents its developed (neural) embedding. 
In practice, the dynamical graph embedding process is stopped whenever the Euclidean distance between the states in successive iterations of \eqref{eq.encoding} goes below a threshold $\varepsilon$, or a maximum number of iterations $\nu$ has been reached.
After initialization, the weights in both 
$\V$ and $\W$ are left untrained, hence the striking efficiency advantage with respect to the approach in \cite{scarselli2009graph}, where stability is not imposed by construction at initialization, 
but rather obtained by a possibly long and costly constrained training process.

The output computation is performed by a readout layer, that linearly combines the representations developed for each vertex of an input graph. The readout computation can be described as follows:
\begin{equation}
\label{eq.readout}
\y(g) = \W_o \, \sum_{v \in V_g}\x(v),
\end{equation}
where $\W_o \in \R^{N_O \times N_H}$ is the readout weight matrix that is adjusted on a set of training samples. Note that the readout operation in \eqref{eq.encoding} includes the application of a global sum-pooling (or aggregation) operator before the linear combination. Training of $\W_o$ is performed in closed-form using Tikhonov regularization as in classical RC approaches \cite{Lukosevicius2009}.

\section{Ring Reservoirs for Graphs}
\label{sec.ringrc}
Given the untrained and randomized approach to the design of network's dynamics, a commonly arising question in the RC field is how to create ``better'' reservoirs than just random ones. To keep the computational advantage of untrained recurrent connections, a possible way to address this question consists in trying to optimize the architecture of the reservoir beforehand. One interesting concept emerged in the literature on dynamical neural models for time-series processing is orthogonality of recurrence matrices. In the temporal domain, this kind of recurrence systems are known to be naturally biased towards having improved memorization skills \cite{white2004short}. Although several ways exist to ``orthogonalize'' a dynamical neural model (see, e.g., \cite{farkavs2016computational,strauss2012design}), a particularly effective approach is to impose a ring-constrained pattern of connectivity among the recurrent neurons in order to connect them to form a cycle.
A number of studies in the context of time-series processing highlighted several advantages of such ring constrained topologies, e.g. in terms of enriched quality of developed representations \cite{tino2019dynamical}, improved memory and predictive performance \cite{rodan2010minimum,strauss2012design}.
For the first time in literature, in this paper we investigate the effects of this type of constrained architectures in the field of neural networks for graph processing. 
We introduce the following two new models that progressively simplify the construction of RC networks for graphs.

\begin{figure*}[htbp]
\centering
\includegraphics[width=1\textwidth]{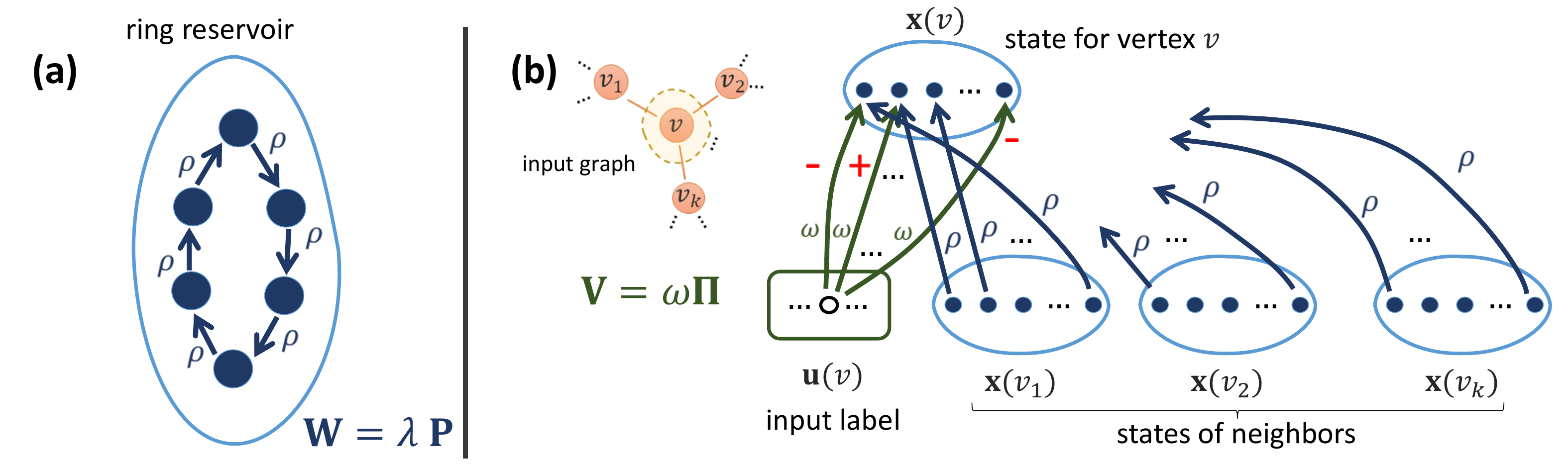}
\caption{MGN: ring-shaped topology of the reservoir (a), and its application to graph encoding (b). The same reservoir topology is used in GRN.}
\label{fig.MGN}
\end{figure*}

\noindent
\textbf{Graph Ring-Reservoir Network (GRN) --}
Connections among reservoir neurons are organized according to a ring topology, where each neuron propagates its activation to the successive one and is fed by the previous one in the cycle. The resulting reservoir weight matrix $\W$ has the shape of a permutation matrix $\mathbf{P}$, where all entries are $0$, except on the subdiagonal and on the top-right corner, i.e.:
\begin{equation}
\label{eq.ring}
\mathbf{P} = \left[
\begin{array}{llll}
0 & 0 & \ldots & 1\\
1 & 0 & \ldots & 0\\
\vdots & \ddots & \ddots & \vdots\\
0 & \ldots & 1 & 0\\
\end{array}
\right], \quad
\W = \lambda \; \mathbf{P}.
\end{equation}
All the non-zero  weights in $\W$ are set to the same value of $\lambda$, which directly controls $\rho(\W)$ and hence can be  used to tune the effective spectral radius of the reservoir system through the relation $\rho = \lambda \, k$. 
The resulting state transition function in \eqref{eq.encoding} is simplified as follows:
\begin{equation}
\label{eq.encodingring}
\X = \tanh(\V \, \U + \lambda \, \mathbf{P} \X \A).
\end{equation}
Notice that 
the effects of stochasticity in the construction of the reservoir are partially removed. The recurrent reservoir weights are indeed 
set in a deterministic way,
and the content of
matrix $\W$ is fully described by just one number (i.e., the spectral radius). 
Randomization enters the network's design solely as regards the setup of the input weight matrix $\V$. 

\noindent
\textbf{Minimal Graph Network (MGN) --}
The reservoir weight matrix $\W$ is ring-shaped as in the case of GRN. In addition,  we impose an architectural simplification also to the input weight matrix $\V$. Specifically, all the input weights are set to the same absolute value $\omega$, while the signs of the weights are chosen deterministically similarly to \cite{rodan2010minimum}, which introduced the idea of ``minimal complexity'' in recurrent architectures.
We construct a sign weight matrix $\mathbf{\Pi}$ of the same size as $\V$. Following the (aperiodic) decimal expansion of the irrational number $\pi$, we row-wise generate the entries in $\mathbf{\Pi}$ using the value $-1$ if the corresponding digit of $\pi$ is smaller than $5$, and the value $+1$ otherwise. 
Accordingly, we have that $\V = \omega \, \mathbf{\Pi}$, and the reservoir state transition function in \eqref{eq.encoding} is further simplified as follows:
\begin{equation}
\label{eq.encodingminimal}
\X = \tanh(\omega \,\mathbf{\Pi}\, \U + \lambda \, \mathbf{P} \X \A).
\end{equation}

Fig.~\ref{fig.MGN} illustrates the simplified reservoir architecture and the state computation in MGN networks. Note that the same ring reservoir topology is used also for GRN.

\noindent
Interestingly, the reservoir of MGN is constructed in a fully deterministic fashion, and any aspect of randomization in its setup has been eliminated. The reservoir system is now completely described by just 2 numbers, i.e., the input scaling parameter $\omega$ and the spectral radius $\rho$. 
The network initialization process is thereby greatly simplified, reducing its degrees of freedom - in comparison to the case of GESN - by a factor that scales with the reservoir size $N_H$.\\
For both GRN and MGN, the readout operation and training is the same as already described for GESN in Section~\ref{sec.rcgraph}.

\section{Experiments}
\label{sec.experiments}
We perform an experimental evaluation of the proposed GRN and MGN models, 
by assessing their predictive performance on several graph classification benchmarks. Our analysis is performed comparatively to both GESN (through our experiments) and state-of-the-art neural networks and kernel methods for graphs (through literature results).

\noindent
\textbf{Datasets.}
We take into consideration 6 graph classification datasets (publicly available online \cite{KKMMN2016}). Two of them, MUTAG \cite{debnath1991structure} and NCI1 \cite{wale2008comparison}, come from the cheminformatics domain, where each input graph is used as representation of a chemical compound: each vertex stands for an atom of the molecule, and edges between vertices represent bonds between atoms. In the case of MUTAG, the dataset contains nitroaromatic compounds and the target classification represents mutagenicity on Salmonella typhimurium. In the case of NCI1, the dataset is relative to anti-cancer screens where the chemicals are assessed as positive or negative to cell lung cancer. For both MUTAG and NCI1, each vertex has an input label representing the corresponding atom type, encoded by a one-hot-encoding scheme into a vector of 0/1 elements.
The other datasets that we use come from the social network analysis domain and are introduced in \cite{yanardag2015deep}. Two of them, i.e., IMDB-BINARY (IMDB-2) and IMDB-MULTI (IMDB-m), are datasets of movie collaboration, where each graph represents the ego network of an actor/actress and the target classification pertains the movie genre (2 for IMDB-2, 3 for IMDB-m). Then, we consider the REDDIT (binary) dataset, where each input graph represents an online discussion thread to be classified in one of 2 possible types of discussions. The last dataset that we use is COLLAB, a collection of graphs representing the ego-networks of researchers, classified according to their areas of research. For all the social network benchmarks, there is no label attached to each vertex in the input graphs, and we use a fixed (uni-dimensional) 1 value as input label for each vertex. 
For all binary classification problems we encoded the target class as a value in $\{-1,+1\}$. For multi-class classification benchmarks each target is represented through a -1/+1 one-hot-encoding of the corresponding class.
Table~\ref{tab.datasets} shows a summary of datasets information.

\begin{table}[tbh]
  \centering
  \begin{tabular}{llllll}
    \toprule
    Dataset & \# graphs & \# vertices (tot) &
    \# vertices (avg) & \#  classes \\ 
    \midrule
    MUTAG       & 188  & 3371   & 17.9  & 2 \\
    IMDB-b      & 1000 & 19773  & 19.8  & 2 \\
    IMDB-m      & 1500 & 19502  & 13.0  & 3 \\
    REDDIT      & 2000 & 859254 & 429.6 & 2 \\
    NCI1        & 4110 & 122747 & 29.9  & 2 \\
    COLLAB      & 5000 & 372474 & 74.5  & 3 \\
    \bottomrule
    \\
  \end{tabular}
 
  \caption{Summary of datasets information.}
  \label{tab.datasets}
\end{table}

\noindent
\textbf{Experimental Settings.}
We conducted experiments with GRN, MGN and GESN. For all the considered RC networks we used dense input weight matrices $\V$
and sparse reservoir weight matrices $\W$. For a fair comparison among the considered RC models, in the case of  GESN we used a sparse pattern of connectivity such that each reservoir neuron is connected to only 1 other reservoir neuron. In this way, the degree of reservoir sparsity is comparable in all the considered settings, with the crucial difference that in \model and GRN such sparsity is not randomized but rather structured following the ring topology described in Section~\ref{sec.ringrc}. 
As regards the stop conditions for the dynamical graph embedding process, we used a convergence threshold of $\varepsilon = 10^{-3}$, and a maximum number of iterations $\nu = 50$.
The hyper-parameters $\omega$ and $\rho$ were explored by random search, i.e., generating a number of $\mathcal{C} = 50$ random reservoir configurations where $\omega$ and $\rho$ were sampled from a uniform distribution on $(0,1)$.
For every reservoir configuration, the Tikhonov regularizer for the readout training was explored in a log-scale grid $\{10^{-10}, 10^{-9}, \ldots, 10^{5}\}$.
We ran experiments considering progressively larger reservoirs, with 
$N_H$ in the range $\{5, 10, 30, 50\}$ for MUTAG, IMDB-2, IMDB-m, and REDDIT, and in the range $\{50, 100, 300, 500\}$ for NCI1 and COLLAB.

To account for randomization in the initialization of the reservoir, for each of the $\mathcal{C}$ reservoir configurations we instantiated a number of $\mathcal{R} = 50$ repetitions (i.e., reservoir guesses), averaging the error on such repetitions. The total number of generated networks is then equal to $\mathcal{C} \times \mathcal{R}$.
Note that for the MGN model the reservoir initialization process is fully deterministic and there is no need to average the performance on multiple repetitions. We thereby considered two possible experimental settings for MGN hyper-parameters search: one ``complete'' and one ``reduced''. In the complete setting we put on par the total effort in the hyper-parameter search, and generated the same total number of reservoir networks as for the other models. In this case, the experiments with MGN span a total number of $\mathcal{C} \times \mathcal{R}$ possible configurations. In the reduced setting, we put on par the number of explored configurations, and limit ourselves to considering a total number of  $\mathcal{C}$ MGN networks. In this case, the 
hyper-parameter search for MGN is $\mathcal{R}$ times faster than those for \model and GESN.

The output class for binary classification tasks was computed by applying the readout equation \eqref{eq.readout}, followed by the sign function to discretize the output in $\{-1,+1\}$.
For multi-class classification tasks, for each graph the output class was assigned in correspondence of the readout unit with the highest activation.
We computed the performance 
in terms of accuracy, following a stratified 10-fold cross validation scheme. The values of the hyper-parameters were chosen on each fold by model selection (individually for each model), using a nested level of stratified 10-fold cross validation.

\begin{figure*}[htbp]
\centering
\includegraphics[width=.32\textwidth]{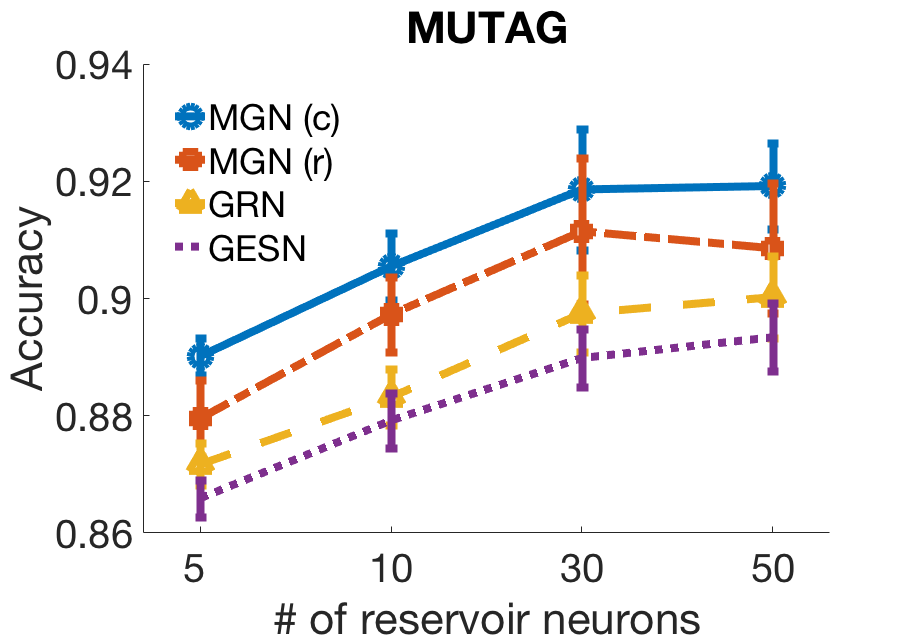}
\includegraphics[width=.32\textwidth]{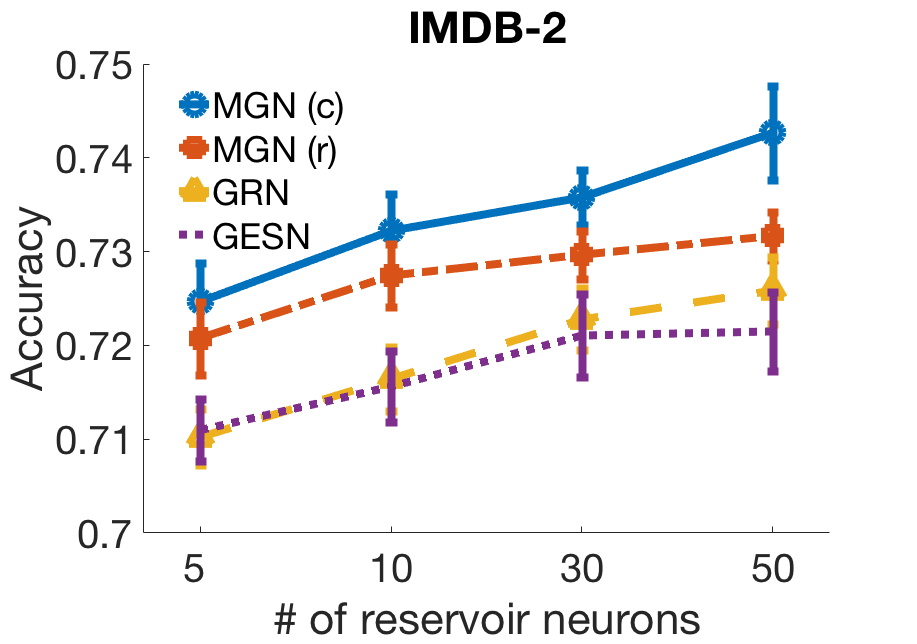}
\includegraphics[width=.32\textwidth]{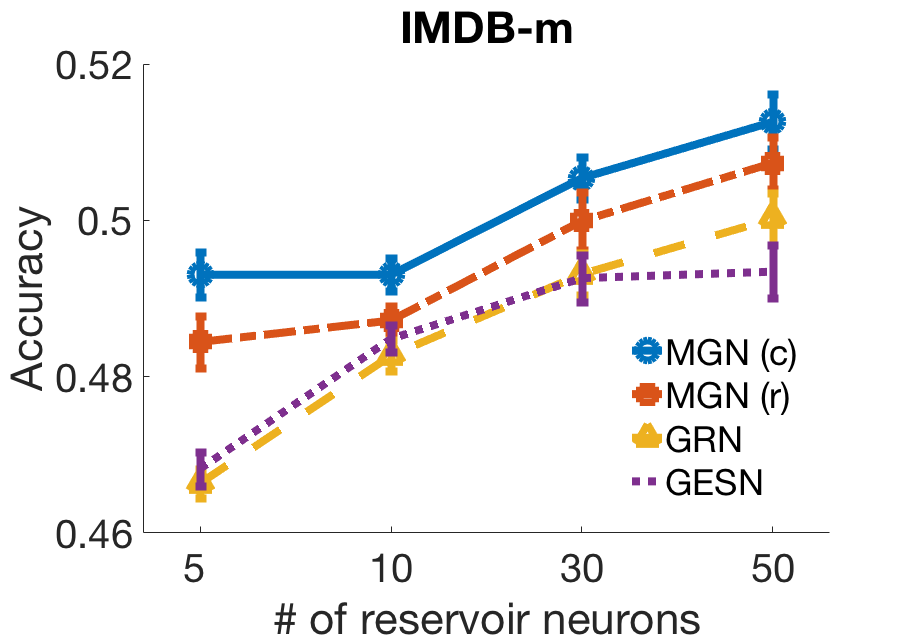}\\
\includegraphics[width=.32\textwidth]{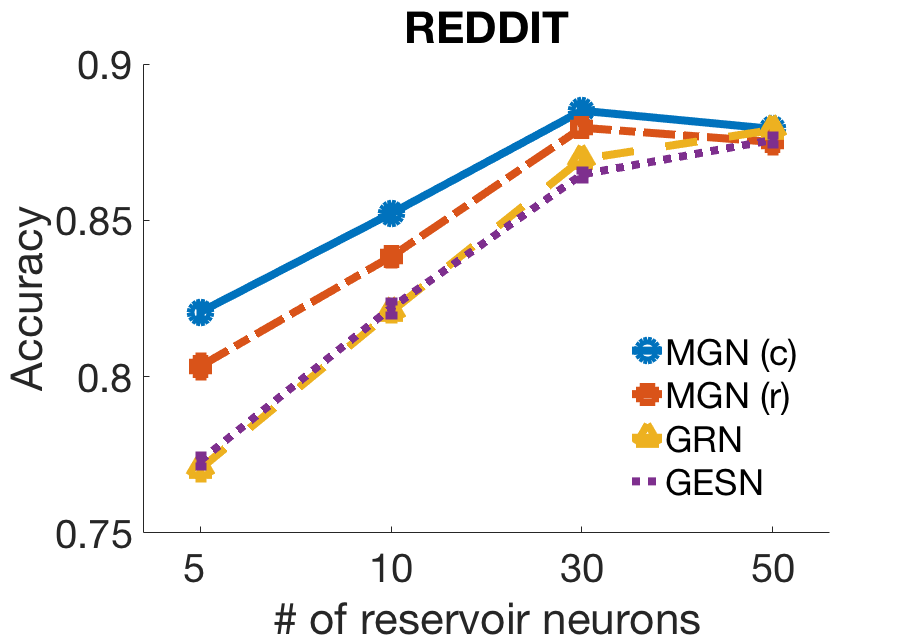}
\includegraphics[width=.32\textwidth]{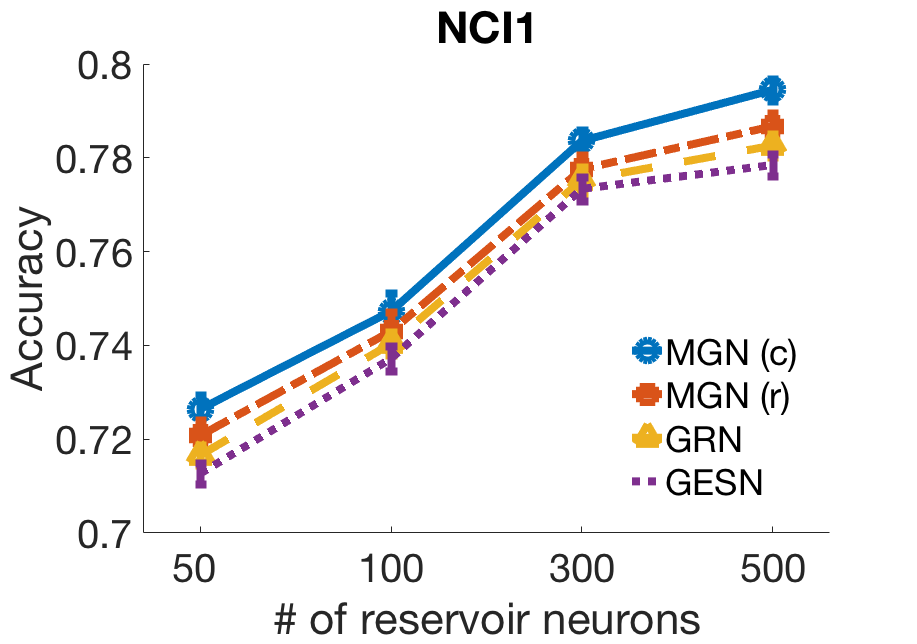}
\includegraphics[width=.32\textwidth]{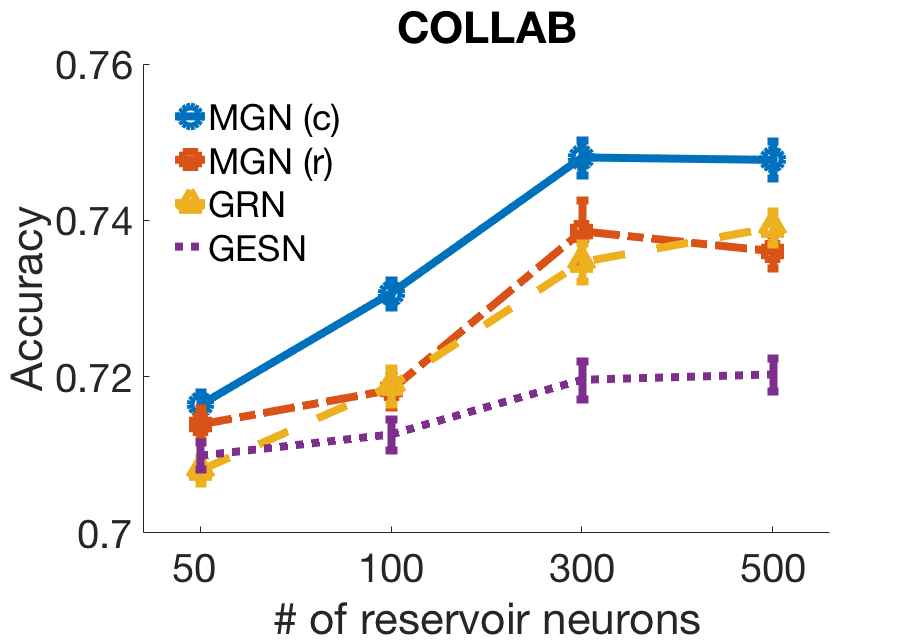}
\caption{Validation accuracy or RC models for graph classification for increasing reservoir size. Results are averaged (and std computed) on the outer 10 folds.}
\label{fig.results}
\end{figure*}

\noindent
\textbf{Results.} 
We first analyzed the behavior of the proposed approaches at the increase of the reservoir network size. To this end, we computed the performance achieved by each model, optimizing the hyper-parameters individually for each possible number of reservoir neurons. The results are shown in Fig.~\ref{fig.results}, in which we report the validation accuracy achieved by \model and MGN (under both complete and reduced settings) in comparison to GESN.
Fig.~\ref{fig.results} indicates a clear general trend, with
both GRN and MGN having higher accuracy than GESN consistently for all reservoir sizes on all tasks.
In particular, the performance gap is evidently wider in the case of MGN. 
Interestingly, despite the architectural simplifications, minimal ring-shaped reservoirs show a higher fitting potential. Already under reduced search settings, MGN has higher performance than GESN and GRN (with almost on par results on COLLAB). Under complete search settings, MGN beats all other models, showing an effective enhancement in the exploration of the RC hyper-parameters space. Remarkably, on 4 over 6 tasks, the performance of the smallest MGN (complete) is in line or even better than the performance of the largest (10 times bigger) GESN.

\begin{table*}[tbh]
  \centering
  \small
  \begin{tabular}{lllllll}
    \toprule
    & MUTAG & IMDB-2 & IMDB-m & REDDIT & NCI1 & COLLAB\\
    \midrule
    MGN \footnotesize{(complete)}  &  
               $87.8_{(\pm 6.3)}$ & 
               $\mathbf{72.7}_{(\pm 3.2)}$ &
               $49.5_{(\pm 2.2)}$ &
               $87.7_{(\pm 1.7)}$ &
               ${78.8}_{(\pm 2.3)}$ &
               ${74.5}_{(\pm 1.4)}$\\
    MGN  \footnotesize{(reduced)}  &  
               $88.3_{(\pm 5.6)}$ & 
               $72.0_{(\pm 3.6)}$ &
               ${50.2}_{(\pm 2.0)}$ &
               $88.1_{(\pm 1.8)}$ &
               $77.6_{(\pm 2.4)}$ &
               ${74.1}_{(\pm 1.3)}$\\ 
    GRN   &
              $88.4_{(\pm 7.6)}$ &
              $71.7_{(\pm 2.8)}$ &
              ${\mathbf{50.5}}_{(\pm 1.4)}$ &
              ${87.6}_{(\pm 1.4)}$ &
              ${78.2}_{(\pm 2.2)}$ &
              ${73.8}_{(\pm 1.4)}$\\
              
    GESN   &
              $88.2_{(\pm 6.3)}$ &
              $71.7_{(\pm 3.6)}$ &
              $48.7_{(\pm 2.1)}$ &
              $87.5_{(\pm 1.1)}$ &
              $77.8_{(\pm 2.0)}$ &
              $72.1_{(\pm 1.6)}$\\
    \midrule
    \vspace{-4mm}
    \\
    FDGNN \cite{gallicchio2020aaai}   &
              $\mathbf{88.5}_{(\pm 3.8)}$ & 
              $72.4_{(\pm 3.6)}$ &
              $50.0_{(\pm 1.3)}$ &
              $\mathbf{89.5}_{(\pm 2.2)}$ &
              $77.8_{(\pm 1.6)}$ &
              $74.4_{(\pm 1.0)}$\\
    DGCNN \cite{zhang2018end}   &
              $85.8_{(\pm 1.7)}$ & 
              $70.0_{(\pm 0.9)}$ &
              $47.8_{(\pm 0.9)}$ &
              $77.1_{(\pm 2.9)}$ & 
              $74.4_{(\pm 0.5)}$ &
              $73.8_{(\pm 0.5)}$\\
    PGC-DGCNN \cite{tran2018filter}   &
              $87.2_{(\pm 1.4)}$ & 
              $71.6_{(\pm 1.2)}$ &
              $47.3_{(\pm 1.4)}$ &
              $-$ & 
              $76.1_{(\pm 0.7)}$ &
              $75.0_{(\pm 0.6)}$\\
    DiffPool \cite{Errica2020A} &
              $-$ & 
              $68.3_{(\pm 6.1)}$ &
              $45.1_{(\pm 3.2)}$ &
              $76.6_{(\pm 2.4)}$ &
              $76.9_{(\pm 1.9)}$ &
              $67.7_{(\pm 1.9)}$\\  
    ECC \cite{Errica2020A} &
              $-$ & 
              $67.8_{(\pm 4.8)}$ &
              $44.8_{(\pm 3.1)}$ &
              $-$ &
              $76.2_{(\pm1.4)}$ &
              $-$\\
    GIN \cite{Errica2020A} &
              $-$ & 
              $66.8_{(\pm 3.9)}$ &
              $42.2_{(\pm 4.6)}$ &
              $87.0_{(\pm 4.4)}$ &
              $80.0_{(\pm1.4)}$ &
              $\mathbf{75.9}_{(\pm 1.9)}$\\
    GraphSAGE \cite{Errica2020A} &
              $-$ & 
              $69.9_{(\pm 4.6)}$ &
              $47.7_{(\pm 3.6)}$ &
              $86.1_{(\pm 2.0)}$ &
              $76.0_{(\pm 1.8)}$ &
              $71.6_{(\pm 1.5)}$\\
    GNN \cite{uwents2011neural} &
              $80.5_{(\pm 0.8)}$ & 
              $-$ &
              $-$ &
              $-$ &
              $-$ &
              $-$\\
    RelNN \cite{uwents2011neural} &
              $87.8_{(\pm 2.5)}$ & 
              $-$ &
              $-$ &
              $-$ &
              $-$ &
              $-$\\
    \midrule
    \vspace{-4mm}
    \\
    GK \cite{yanardag2015deep} &
              $81.4_{(\pm 1.7)}$ & 
              $65.9_{(\pm 1.0)}$ &
              $43.9_{(\pm 0.4)}$ &
              $77.3_{(\pm 0.2)}$ &
              $62.5_{(\pm 0.3)}$ & 
              $72.8_{(\pm 0.6)}$\\
    DGK \cite{yanardag2015deep} &
              $82.7_{(\pm 1.5)}$ & 
              $67.0_{(\pm 0.6)}$ &
              $44.6_{(\pm 0.5)}$ &
              $78.0_{(\pm 0.4)}$ &
              $62.5_{(\pm 0.3)}$ & 
              $73.1_{(\pm 0.3)}$\\
    WL \cite{zhang2018end} &
              $84.1_{(\pm 1.9)}$ & 
              $-$ &
              $-$ &
              $-$ &
              $\mathbf{84.5}_{(\pm 0.5)}$ & 
              $-$\\
    PK \cite{zhang2018end} &
              $76.0_{(\pm 2.7)}$ & 
              $-$ &
              $-$ &
              $-$ &
              $82.5_{(\pm0.5)}$ & 
              $-$\\
    \bottomrule
    \\
    \vspace{-5mm}
  \end{tabular}
  \caption{Test accuracy of RC models for graph classification (compared to literature methods). Results are averaged (and std computed) on the outer 10 folds. 
  The best overall accuracy on each task is highlighted in bold font.}
  \label{tab.results}
\end{table*}

We then analyze the overall generalization performance of the proposed architectures by individually optimizing (through model selection) the hyper-parameters of each RC model, taking collectively into account all the possible reservoir sizes. The achieved test accuracies are reported in Tab.~\ref{tab.results}. In the same table, we
report the performance achieved by a significant selection of state-of-the-art ML models for graph classification from the literature. 
In particular, we considered the following graph neural networks:
Fast and Deep Graph Neural Network (FDGNN) \cite{gallicchio2020aaai}, a recently introduced deep RC approach for graphs, Deep Graph Convolutional Neural Network (DGCNN) \cite{zhang2018end}, Parametric Graph Convolution DGCNN (PGC-DGCNN) \cite{tran2018filter}, DiffPool \cite{ying2018hierarchical},
Edge-Conditioned Convolution network (ECC) \cite{simonovsky2017dynamic}, Graph Isomorphism Network (GIN) \cite{xu2018powerful} GraphSAGE \cite{hamilton2017inductive}, Graph Neural Network \cite{scarselli2009graph} and Relational Neural Network \cite{blockeel2003aggregation}. We also considered a number of kernel methods for graphs, including Graphlet Kernel (GK) \cite{shervashidze2009efficient}, Deep GK (DGK), Weisfeiler-Lehman Kernel (WL) \cite{shervashidze2011weisfeiler}, and Propagation Kernel (PK) \cite{neumann2016propagation}.
Results for these models are quoted from the literature references indicated in the table, where the experimental settings for model selection were as close as possible to be rigorous as ours (the only exception is the result of DGCNN on REDDIT, which is taken from \cite{Errica2020A}).

Tab.~\ref{tab.results} confirms the already observed general trend of ring-constrained RC networks variants beating standard GESN. 
Specifically, both GRN and MGN (in each of the search settings) outperform GESN on 
5 out of 6 tasks.
A remarkable case is given by the largest dataset, i.e., COLLAB, where the performance boost is particularly evident (up to more than $2\%$ of accuracy). We find it particularly intriguing that the observed accuracy improvement is obtained by just by imposing a specific (and rather simple) structure to the architectural setup of the network. In the case of MGN this architectural effect is combined with the already noted ability to search the hyper-parameters space more extensively.
Results of ring reservoir networks compare well also with literature results. In this regard, notice that while literature works on deep learning for graphs often imply complex end-to-end fully trained multi-layered architectures with several thousands of trainable weights, our results here were achieved with very simple (single-layered) recursive untrained architectures, and
entail a maximum number of trained weights of 501 for NCI1 and COLLAB, and of 51 for all the other tasks.
Despite this, our proposed models establish new sate-of-the-art results on IMDB-2 and IMDB-m. On MUTAG, REDDIT and COLLAB our achieved performances are very close to the highest ones in literature, while on NCI1 - where WL and PK kernels score best - the obtained accuracy is not far from the best performing neural network (GIN).    
Finally, looking also at the accuracy of FDGNN, we find it interesting to point out that on 4 out of 6 tasks, the highest accuracy is achieved by a model in the RC class (either in a shallow, or deep architectural setting).

\section{Conclusions}
\label{sec.conclusions}
We have studied simple RC architectures for graph processing where the hidden recurrent neurons are constrained to a ring topology. The introduced simplifications allowed us to progressively reduce, up to eliminate, the aspects of stochasticity in the initialization of reservoir 
networks for graphs. Our experimental analysis indicated 
that such ``minimal'' RC architectures 
enable an effective exploration of the hyper-parameters, often leading to an improved performance. 
The provided results also
pointed out that  despite their simplicity, ring-reservoir neural networks are particularly effective in graph classification benchmarks, reaching a performance that is comparable to (and sometimes even better than) that of several more complex ML models for graphs.

The evidences illustrated in this paper naturally support future research investigations, aiming at extending the observed advantages of constrained reservoir architectures both to the construction of more effective deep RC models and to the design of more efficient learning algorithms for trained neural networks for graphs.

\bibliography{references.bib}
\bibliographystyle{IEEEtran}

\end{document}